\documentclass[letterpaper]{article} % DO NOT CHANGE THIS
\makeatletter
\def\input@path{{AAAI27-QuantaFlow/}}
\makeatother
\usepackage{aaai2027}  % DO NOT CHANGE THIS
\usepackage[hyphens]{url}  % DO NOT CHANGE THIS
\usepackage{graphicx} % DO NOT CHANGE THIS
\usepackage{natbib}  % DO NOT CHANGE THIS AND DO NOT ADD ANY OPTIONS TO IT
\usepackage{caption} % DO NOT CHANGE THIS AND DO NOT ADD ANY OPTIONS TO IT
\usepackage{amsmath}
\usepackage{amssymb}

\DeclareCaptionStyle{ruled}{labelfont=normalfont,labelsep=colon,strut=off} % DO NOT CHANGE THIS

\usepackage{booktabs}

\title{Optical Flow from Photons}
\author{
    Wendi Liu,
    Weichao Zeng,
    Weihang Ran,
    Yujie Lu,
    Yinqiang Zheng\corresponding
}
\affiliations{
    The University of Tokyo\\
}

\nocopyright

\begin{document}

\maketitle

\begin{abstract}
Optical flow remains challenging in high-speed and low-light scenes, where the limited frame rate and sensitivity of conventional cameras lead to motion blur and underexposure. Single-photon avalanche diode (SPAD) cameras offer single-photon sensitivity and extremely fine temporal sampling. However, individual slices in these high FPS binary photon streams are too sparse for dense correspondence. Temporal aggregation can provide the spatial cues required by optical flow, but accumulating photons at fixed coordinates blurs moving structures. Motion-aware aggregation can reduce this blur, yet it depends on the flow being estimated. To address this dependency, we propose QuantaFlow, the first method for dense optical flow directly from SPAD streams. Instead of constructing a fixed input representation, QuantaFlow embeds SPAD representation construction into iterative flow refinement. At each iteration, the current flow coarsely aligns the slices within the source and target sub-streams. A photon-flux transformation then constructs multi-scale representations containing intensity and structural cues, while adaptive multi-scale fusion balances photon noise and residual motion blur at each pixel. The fused representations drive a feature-warping flow update, and the refined flow guides representation construction in the next iteration. We further construct a synthetic dataset for SPAD optical-flow training and evaluation. Experiments on the synthetic dataset and real-world SPAD data demonstrate the effectiveness and generalization of QuantaFlow.
\end{abstract}


\section{Introduction}

Optical flow is a fundamental problem in computer vision and provides motion information for applications such as video interpolation~\cite{Jiang_2018_CVPR, bao2019depth}, motion segmentation~\cite{narayana2013coherent, tokmakov2017learning}, autonomous driving~\cite{capito2020optical, wang2021end}, and robot navigation~\cite{mcguire2017efficient}. Despite substantial progress on standard RGB benchmarks, estimating flow remains difficult in high-speed and low-light scenes, where the limited frame rate and sensitivity of conventional cameras lead to motion blur, underexposure, and the loss of structures required for correspondence. The emergence of neuromorphic sensors has enabled optical flow to be estimated from new visual modalities, notably event~\cite{gehrig2024, gehrig2021eraft} and spike streams~\cite{hu2022scflow,zhao2024hstsf,zhao2026scflow}. However, event cameras record brightness changes rather than absolute scene intensity and therefore provide little signal in static or weak-contrast regions~\cite{gallego2022event}. Spike cameras integrate photocurrent until a firing threshold is reached; under low illumination, the resulting long inter-spike intervals leave less temporal information for resolving fast motion. SPAD cameras offer a different sensing mechanism in which a photon-triggered avalanche enables single-photon detection with negligible read noise and extremely fine temporal sampling~\cite{bruschini2019spad,seets2021motion}. These properties suggest strong potential for optical flow in fast and photon-limited scenes.
\begin{figure}[t]
    \centering
    \includegraphics[width=\columnwidth]{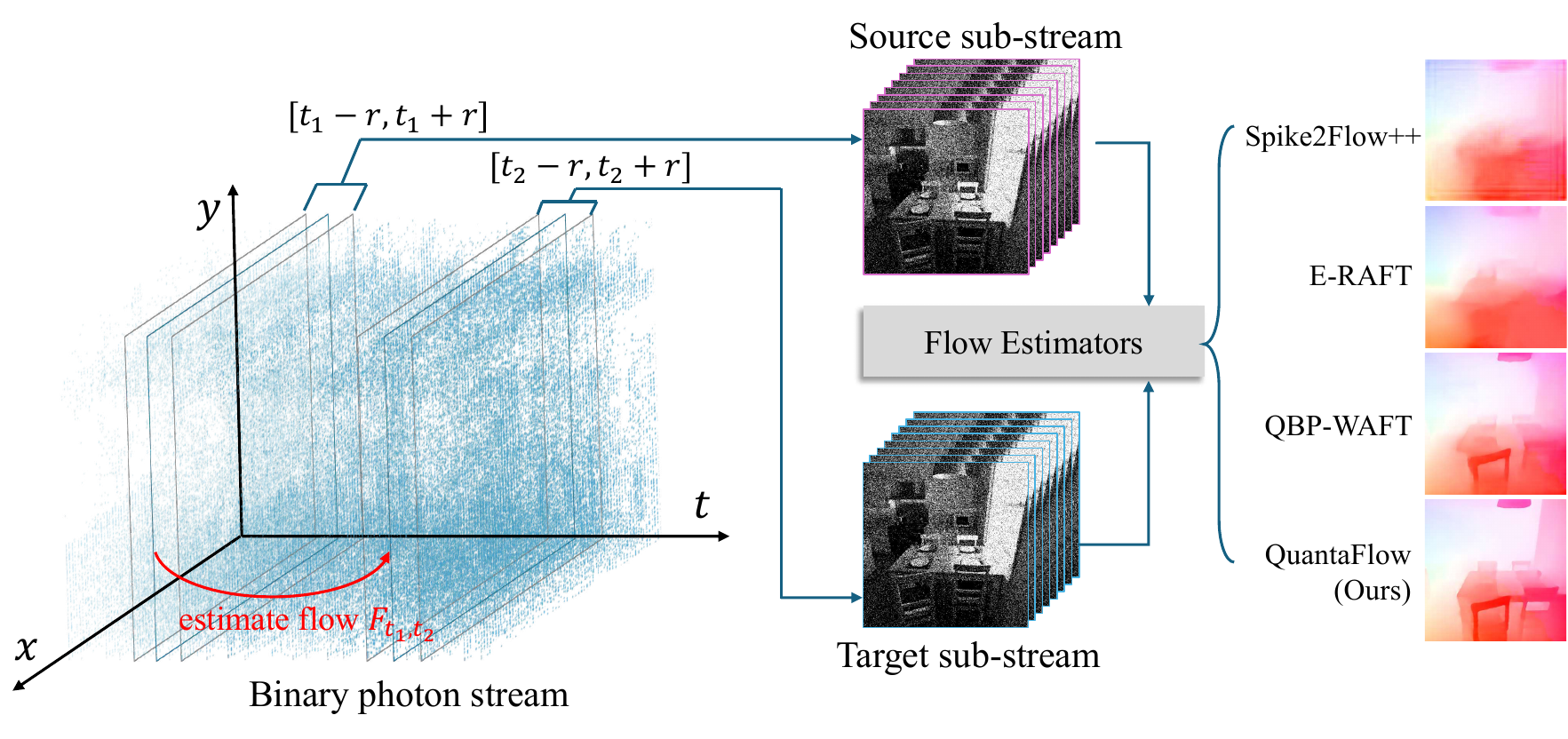}
    \caption{Optical flow estimation from SPAD photon streams. A SPAD sensor records a high-rate sequence of binary photon-detection slices. The '1' means at least one photon is detected at the pixel, and the '0' means no photon is detected. Given two sub-streams centered at $t_1$ and $t_2$, the task is to estimate $F_{t_1,t_2}$. QuantaFlow recovers more complete object motion and sharper boundaries than frame-, spike-, and event-based methods.}
    \label{fig:intro}
\end{figure}

Despite these advantages, dense optical flow estimation directly from SPAD streams remains unexplored. The key challenge is to convert sparse binary photon detections into representations that preserve sufficient spatial structure for correspondence. An individual SPAD frame contains very few detections and reveals almost no continuous texture, so existing RGB flow networks cannot match two such frames directly. Summing nearby frames recovers spatial structure, but a fixed-pixel sum assumes that the scene remains stationary during the accumulation. Under camera or scene motion, photons from different surfaces are mixed, with the largest errors occurring near depth discontinuities and image boundaries. Without motion information, temporal aggregation inevitably blurs the spatial structures of moving scenes. Aggregation thus requires the very motion that flow estimation is meant to recover.

To resolve this interdependence, we propose QuantaFlow, which refines the SPAD representation together with optical flow. At each iteration, Motion-Guided Alignment (MGA) uses the current flow to coarsely align the slices within the source and target sub-streams, reducing motion blur before photon aggregation. Photon-Flux Transformation (PFT) then constructs representations with intensity and spatial structure at multiple temporal scales. Adaptive Multi-Scale Fusion (AMF) combines these representations at each pixel to balance photon noise and residual motion blur. The fused representations are used to update the flow through feature-level warping, and the refined flow returns to MGA to reconstruct sharper representations in the next iteration. We refer to this coupled refinement as Iterative Representation Update (IRU).

Our contributions are as follows:
\begin{itemize}
    \item We introduce the first method for dense optical flow directly from SPAD photon streams, targeting high-speed and photon-limited scenes.
    \item We develop a motion-aware multi-scale SPAD representation that combines slice-level alignment, photon-flux transformation, and adaptive fusion. Its iterative update jointly improves the representation and optical flow.
    \item We construct a synthetic dataset for training and evaluation. Experiments on the synthetic dataset and real-world SPAD data demonstrate the effectiveness of QuantaFlow.
\end{itemize}

\section{Related Work}

\subsection{Optical Flow from RGB Images}

Classical methods cast optical flow as a variational energy that balances brightness constancy against spatial smoothness~\cite{horn1981}. Learning-based estimators replaced this optimization with end-to-end networks, starting from FlowNet~\cite{dosovitskiy2015flownet} and coarse-to-fine designs such as PWC-Net~\cite{sun2018pwcnet}. RAFT~\cite{raft2020} introduced recurrent refinement over a 4D correlation volume and became the dominant paradigm. Later work improved accuracy with global motion aggregation, transformers, and global matching~\cite{jiang2021gma,huang2022flowformer,shi2023flowformerpp,xu2022gmflow}, and with improved training and uncertainty modeling in SEA-RAFT~\cite{searaft2024}. WAFT~\cite{waft2026} showed that high-resolution warping can replace the cost volume, trading explicit correlation for memory efficiency. These methods are designed for RGB imagery and assume sharp and well-exposed frames, which high-speed or low-light SPAD capture does not provide.

\subsection{Optical Flow from Neuromorphic Sensors}

Neuromorphic and high-speed sensors have prompted optical-flow estimators tailored to their output modalities. Event cameras report asynchronous brightness changes, and dedicated methods estimate flow from event tensors~\cite{zhu2018evflownet,gehrig2021eraft} or directly as a continuous-time field~\cite{gehrig2024}. Spike cameras instead encode incident light through an integrate-and-fire mechanism. SCFlow~\cite{hu2022scflow} introduced a spike-specific flow network with a motion-guided temporal representation. Subsequent methods jointly reconstruct intensity and estimate flow~\cite{chen2023spikeflow}, fuse spike features across spatial and temporal scales~\cite{zhao2024hstsf}, and estimate flow from continuous spike streams~\cite{zhao2026scflow,zhao2022learning}. These works demonstrate that high-rate binary sensor streams require representations tailored to both the sensing mechanism and the flow task. SPAD cameras follow a different image-formation process, and optical flow directly from SPAD streams remains unexplored.

\subsection{SPAD-Based Imaging}
Passive SPAD imaging research has focused on reconstructing high-quality
intensity images or video from binary photon streams. Quanta burst
photography~\cite{ma2020qbp} aligns and merges binary frames to suppress
noise and motion blur, and subsequent works extend this
reconstruct-first paradigm to dynamic scenes and video
restoration~\cite{chennuri2024quanta}. Motion is also exploited to adapt
the integration window~\cite{seets2021motion}, and SoDaCam~\cite{sundar2023sodacam} re-purposes the photon cube to emulate other sensor modalities. In these works, motion only serves image reconstruction; to the best of our knowledge, no existing work estimates dense optical flow directly from SPAD streams.

\begin{figure*}[t]
\centering
\includegraphics[width=\textwidth]{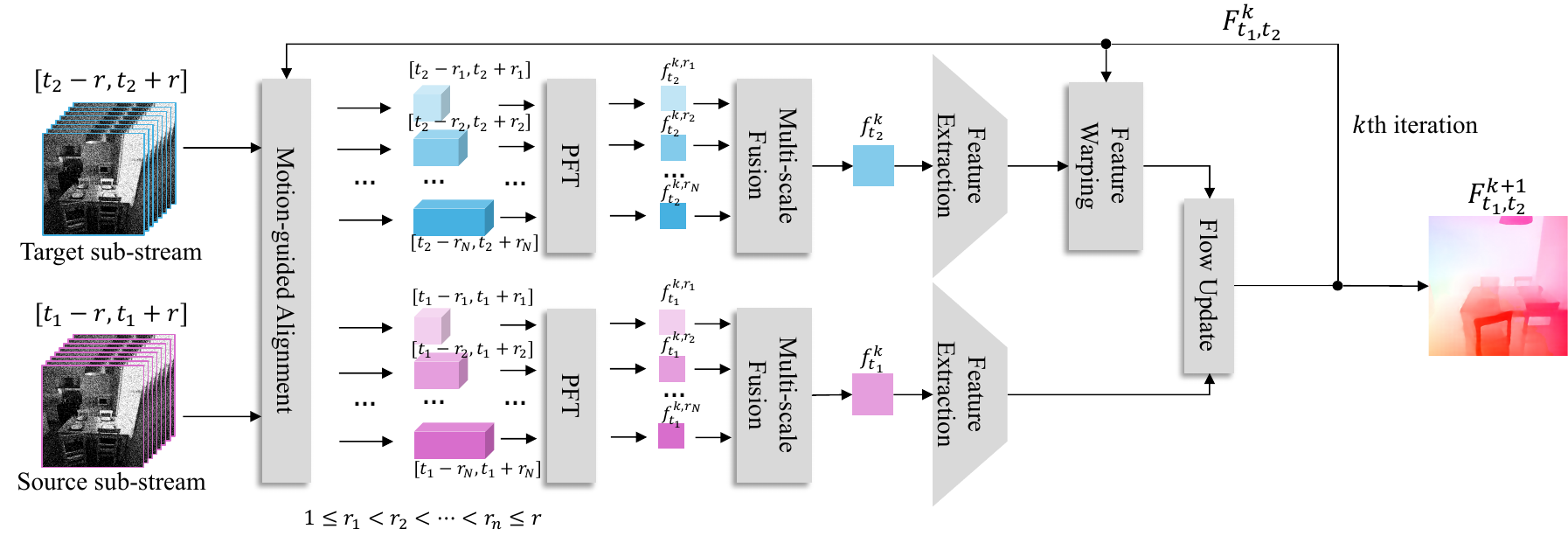}
\caption{Overview of QuantaFlow. The current flow estimate first guides slice alignment within the source and target SPAD sub-streams. The Photon-Flux Transformation (PFT) module constructs a SPAD representation at each temporal scale, and the Adaptive Multi-Scale Fusion (AMF) module combines them for feature extraction and flow refinement. The refined flow is then fed back to improve slice alignment in the next iteration.}
\label{fig:framework}
\end{figure*}

\section{Preliminary}

\subsection{SPAD Imaging Model}

A single-photon avalanche diode (SPAD) sensor records whether photons arrive at each pixel during a short exposure. It outputs $1$ when at least one photon is detected and $0$ otherwise, producing a binary stream $S_n\in\{0,1\}^{W\times H}$ indexed by time $n$.

Let $I_n(u)$ denote the linear scene intensity at pixel $u$. We denote the latent photon flux by $H_n(u)=\alpha I_n(u)$ and the corresponding photon detection probability by $p_n(u)=1-e^{-H_n(u)}$. Following the SPAD image formation model~\cite{garg2026gqir}, the binary observation follows the Bernoulli distribution:
\begin{equation}
    S_n(u)
    \sim
    \mathrm{Bernoulli}\!\left(1-e^{-\alpha I_n(u)}\right).
    \label{eq:bernoulli}
\end{equation}
The factor $\alpha$ controls the expected photon-per-pixel level. Eq.~\eqref{eq:bernoulli} shows that the photon detection probability $p_n(u)$ is a nonlinear and saturating response to the latent flux $H_n(u)$.

At temporal scale $r_i$, a window centered at $t_e$ contains $2r_i+1$ binary slices. When the photon flux remains approximately constant within this window, the accumulated detections follow
\begin{equation}
    D_{t_e}^{r_i}(u)
    =
    \sum_{\delta=-r_i}^{r_i}S_{t_e+\delta}(u)
    \sim
    \mathrm{Binomial}\!\left(2r_i+1,p_{t_e}(u)\right).
    \label{eq:binomial}
\end{equation}
The normalized count $p_{t_e}^{r_i}(u)=D_{t_e}^{r_i}(u)/(2r_i+1)$ estimates the photon detection probability, with
\begin{equation}
    \mathrm{Var}\!\left[p_{t_e}^{r_i}(u)\right]
    =
    \frac{p_{t_e}(u)\left(1-p_{t_e}(u)\right)}{2r_i+1}.
    \label{eq:variance}
\end{equation}
The variance decreases as the temporal scale grows, leading to the improved photon statistics of longer windows.

\subsection{Problem Statement}

Given two timestamps $t_1$ and $t_2$ separated by $\Delta t=t_2-t_1$, let $F_{t_1,t_2}$ denote the optical flow that maps each pixel at $t_1$ to its corresponding position at $t_2$. The input consists of two SPAD sub-streams $S^1=\{S_n\}_{n=t_1-r}^{t_1+r}$ and $S^2=\{S_n\}_{n=t_2-r}^{t_2+r}$ centered at the two timestamps, each containing $2r+1$ binary slices. QuantaFlow directly estimates $F_{t_1,t_2}$ from this pair of sub-streams.

The high sampling rate of SPAD sensors yields very small displacements between adjacent slices. Following prior spike-flow protocols~\cite{hu2022scflow}, we set $\Delta t\in\{10,20\}$ to form small- and large-displacement settings for optical flow estimation.

\section{Method}

\subsection{Overview}
Given the two SPAD sub-streams centered at $t_1$ and $t_2$, QuantaFlow estimates the forward flow through $K$ iterations. As shown in Fig.~\ref{fig:framework}, the $k$-th iteration starts from the current estimate $F^k_{t_1,t_2}$. Motion-Guided Alignment (MGA) uses this flow to coarsely align every slice in each sub-stream to its central slice. Photon-Flux Transformation (PFT) converts the aligned sub-streams at different temporal scales into representations $f^{k,r_i}_{t_1}$ and $f^{k,r_i}_{t_2}$ that encode intensity and spatial structure. Adaptive Multi-Scale Fusion (AMF) further combines these multi-scale representations into $f^k_{t_1}$ and $f^k_{t_2}$. Referring to WAFT~\cite{waft2026}, the flow update module extracts features from the fused representations, warps them at the feature level using $F^k_{t_1,t_2}$, and applies an update unit to estimate $F^{k+1}_{t_1,t_2}$. The updated flow then guides the representation construction in the next iteration. We refer to this process as Iterative Representation Update (IRU) and initialize the flow with $F^0_{t_1,t_2}=0$.

This iterative construction couples the SPAD representation with optical flow. Better motion estimates yield more coherent photon aggregation, while sharper SPAD representations provide stronger structures for correspondence in the following update.

\subsection{Motion-Guided Alignment}

Optical flow estimation relies on stable spatial features that can be matched between two timestamps. For the binary SPAD photon stream with the three-dimensional spatiotemporal structure, these features must be constructed by aggregating photon detections along the temporal dimension. Direct aggregation at fixed sensor coordinates, however, mixes detections from different scene points under motion and blurs the resulting spatial structures. Therefore, we need to align the slices within each temporal window along the motion trajectory to its reference timestamp before constructing the SPAD representation at every iteration.

The extremely high sampling rate of a SPAD sensor makes motion within a short temporal window temporally oversampled. We can assume that the variations of the scene point displacement between adjacent slices are stable and uniform. Under this assumption, the current flow estimate can be temporally scaled to approximate the displacement of every slice relative to the reference timestamp, providing a coarse alignment that reduces aggregation blur. At iteration $k$, $F^k_{t_1,t_2}$ represents the displacement over $\Delta t=t_2-t_1$. For a slice at $t_e+\delta$, its displacement relative to $t_e$ is approximated by
\begin{equation}
    d^{k,e}_{\delta}(u)
    =
    \frac{\delta}{\Delta t}F^k_{t_1,t_2}(u),
    \qquad e\in\{1,2\}.
    \label{eq:guided_displacement}
\end{equation}
This approximation allows the current flow estimate to provide a coarse motion trajectory for the photon detections surrounding both timestamps. Each slice is then coarsely aligned to its center as
\begin{equation}
    \widetilde{S}^{k,e}_{\delta}(u)
    =
    S_{t_e+\delta}\!\left(u+d^{k,e}_{\delta}(u)\right).
    \label{eq:guided_slice}
\end{equation}
MGA turns fixed-pixel accumulation into photon aggregation along the estimated motion trajectory. The aligned detections are more likely to originate from the same scene point, reducing motion blur and preserving sharper spatial structures for the subsequent representation construction.

\subsection{Photon-Flux Transformation}

Constructing the spatial representation from the SPAD stream requires an appropriate temporal scale. A short window preserves sharp spatial structures but contains few photon detections, resulting in pronounced quantum fluctuations and Poisson noise. A long window collects more photons and improves the signal-to-noise ratio, but tends to blur spatial structures. A single temporal scale therefore cannot provide sufficient information under different noise and motion conditions. Accordingly, we extract $N$ aligned sub-streams at different temporal scales and recover intensity and spatial information from each of them.

At temporal scale $r_i$, we use $2r_i+1$ slices centered at $t_e$. The most direct way to obtain an intensity response is to accumulate the aligned photon detections. The normalized accumulation is
\begin{equation}
    p_{t_e}^{k,r_i}(u)
    =
    \frac{1}{2r_i+1}
    \sum_{\delta=-r_i}^{r_i}
    \widetilde{S}^{k,e}_{\delta}(u).
    \label{eq:scale_aggregation}
\end{equation}
This quantity estimates the photon detection probability. As shown in Eq.~\eqref{eq:bernoulli}, $p_{t_e}^{k,r_i}(u)$ is a nonlinear response to the latent photon flux. We therefore invert this response to obtain
\begin{equation}
    H_{t_e}^{k,r_i}(u)
    =
    -\log\!\left(1-p_{t_e}^{k,r_i}(u)\right),
    \label{eq:pft}
\end{equation}
where $H_{t_e}^{k,r_i}$ provides an estimate of the latent photon flux and reveals the underlying scene intensity.

Eq.~\eqref{eq:variance} further shows that the reliability of this estimate depends on both the photon detection probability and the number of observations. When the temporal window is short and photon detections are sparse, the relative uncertainty of $p_{t_e}^{k,r_i}$ is high, and direct nonlinear inversion propagates these fluctuations into $H_{t_e}^{k,r_i}$. We therefore introduce an intensity estimator $\mathcal{P}_{\theta}$ whose weights are shared across temporal scales. Implemented as a lightweight residual network, it jointly considers the detection probability, the recovered flux, and the temporal scale to adaptively extract the latent spatial structure,
\begin{equation}
    f^{k,r_i}_{t_e}
    =
    \mathcal{P}_{\theta}\!\left(
    p_{t_e}^{k,r_i},
    H_{t_e}^{k,r_i},
    r_i
    \right).
    \label{eq:pft_adapter}
\end{equation}
The same estimator is also shared by the two sub-streams. We set $N=4$ and use $r_i=5,15,25,35$. The settings $r_i=5$ and $r_i=35$ correspond to the shortest and longest temporal windows, respectively.

\subsection{Adaptive Multi-Scale Fusion}

PFT produces $N$ SPAD representations $\{f^{k,r_i}_{t_e}\}_{i=1}^{N}$ at different temporal scales. AMF further estimates the pixel-wise weight for every scale from the local scene information and combines the scale-specific representations through weighted fusion,
\begin{equation}
    \begin{aligned}
    w_{t_e}^{k,r_i}(u)
    &=
    \underset{i}{\operatorname{softmax}}\!\left(
    \mathcal{A}_{\phi}\!\left(
    \left\{f^{k,r_j}_{t_e}\right\}_{j=1}^{N}
    \right)(u)
    \right),\\
    f^k_{t_e}(u)
    &=
    \sum_{i=1}^{N}
    w_{t_e}^{k,r_i}(u)f^{k,r_i}_{t_e}(u),
    \end{aligned}
    \label{eq:adaptive_fusion}
\end{equation}
where $\mathcal{A}_{\phi}$ denotes the adaptive weighting module.

At motion boundaries or in regions with high photon flux, AMF assigns larger weights to short-scale representations to preserve sharp spatial structures. In regions with smooth motion or low photon flux, it favors long-scale representations to collect sufficient photon information. The fused representation therefore adapts the trade-off between signal-to-noise ratio and motion blur to the local scene, making fuller use of the spatiotemporal information than selecting a single scale or uniformly averaging all scales.

\begin{table*}[t]
    \centering
    \resizebox{\textwidth}{!}{%
    \begin{tabular}{lcccccccccc}
    \toprule
    Method & \multicolumn{5}{c}{$\Delta t=10$} & \multicolumn{5}{c}{$\Delta t=20$} \\
    \cmidrule(lr){2-6}\cmidrule(lr){7-11}
     & EPE $\downarrow$ & AE $\downarrow$ & 1PE $\downarrow$ & 2PE $\downarrow$ & 3PE $\downarrow$ & EPE $\downarrow$ & AE $\downarrow$ & 1PE $\downarrow$ & 2PE $\downarrow$ & 3PE $\downarrow$ \\
    \midrule
    \multicolumn{11}{l}{\textbf{RGB reference}} \\
    RAFT (RGB) & 0.9762 & 5.6554 & 0.2591 & 0.1137 & 0.0638 & 2.1129 & 5.3123 & 0.4618 & 0.2592 & 0.1745 \\
    WAFT (RGB) & 0.7518 & 4.2848 & 0.1817 & 0.0916 & 0.0525 & 1.6037 & 3.9249 & 0.3194 & 0.1830 & 0.1296 \\
    \midrule
    \multicolumn{11}{l}{\textbf{Frame-based methods}} \\
    RAFT-r5 & 1.8575 & 9.9556 & 0.4968 & 0.2707 & 0.1745 & 3.4137 & 7.9684 & 0.6776 & 0.4353 & 0.3002 \\
    RAFT-r35 & 2.1998 & 8.6368 & 0.5608 & 0.3147 & 0.1981 & 4.2404 & 7.8761 & 0.7687 & 0.5375 & 0.3716 \\
    WAFT-r5 & 1.4332 & 6.5454 & 0.3844 & 0.1842 & 0.1131 & 2.9804 & 5.7378 & 0.5554 & 0.3366 & 0.2367 \\
    WAFT-r35 & 1.7155 & 6.8584 & 0.4474 & 0.2213 & 0.1330 & 3.5577 & 6.1849 & 0.6651 & 0.4461 & 0.3052 \\
    QBP-WAFT & 1.3902 & 7.1670 & 0.4185 & 0.1884 & 0.0973 & 2.6882 & 5.9406 & 0.5716 & 0.3443 & 0.2345 \\
    \midrule
    \multicolumn{11}{l}{\textbf{Spike/Event-based methods}} \\
    E-RAFT & 1.9351 & 10.5408 & 0.5605 & 0.2857 & 0.1727 & 4.2389 & 10.7590 & 0.7716 & 0.5634 & 0.4155 \\
    HiST-SFlow & 2.5712 & 13.4574 & 0.6127 & 0.3922 & 0.2678 & 4.4815 & 11.3206 & 0.7864 & 0.5739 & 0.4318 \\
    SCFlow & 4.4281 & 28.8328 & 0.7968 & 0.6090 & 0.4789 & 9.1376 & 28.2913 & 0.9112 & 0.7877 & 0.6802 \\
    Spike2Flow++ & 2.2212 & 10.7403 & 0.5794 & 0.3290 & 0.2057 & 4.5852 & 10.5538 & 0.7892 & 0.5858 & 0.4432 \\
    \midrule
    \multicolumn{11}{l}{\textbf{Ours}} \\
    QuantaFlow & \textbf{1.1502} & \textbf{5.3361} & \textbf{0.2867} & \textbf{0.1515} & \textbf{0.0940} & \textbf{2.6096} & \textbf{4.9139} & \textbf{0.4344} & \textbf{0.2830} & \textbf{0.2111} \\
    \bottomrule
    \end{tabular}
    }
    \caption{Main comparison on the simulated SPAD optical-flow benchmark. Results are reported separately for the $\Delta t=10$ and $\Delta t=20$ settings. The suffix -r$x$ denotes accumulation over $2x+1$ SPAD frames centered at $t_1$ and $t_2$. RAFT (RGB) and WAFT (RGB) take ground-truth RGB frames as input and serve as reference upper bounds.}
    \label{tab:main_results}
\end{table*}

\subsection{Iterative Representation Update}

For optical-flow update, we follow the warp-and-refine design of WAFT~\cite{waft2026}. A shared encoder extracts dense features $g^k_{t_1}$ and $g^k_{t_2}$ from the fused source and target SPAD representations $f^k_{t_1}$ and $f^k_{t_2}$, respectively. Given the current flow $F^k_{t_1,t_2}$, feature warping fetches the target feature corresponding to each source pixel,
\begin{equation}
    \widetilde{g}^k_{t_2}(u)
    =
    g^k_{t_2}\!\left(u+F^k_{t_1,t_2}(u)\right).
    \label{eq:feature_warp}
\end{equation}
We use a DPT-based~\cite{dpt2021} recurrent update unit $\mathcal U$ to process the source feature, the warped target feature, and the current hidden state. It predicts a residual flow update and refines the current estimate as
\begin{equation}
    \begin{aligned}
    \left(\Delta F^k_{t_1,t_2},h^{k+1}\right)
    &=\mathcal U\!\left(
    g^k_{t_1},\widetilde{g}^k_{t_2},h^k
    \right),\\
    F^{k+1}_{t_1,t_2}
    &=F^k_{t_1,t_2}+\Delta F^k_{t_1,t_2}.
    \end{aligned}
    \label{eq:flow_update}
\end{equation}

Different from WAFT, the current flow in our method QuantaFlow is used not only for feature-level warping but also for slice-level alignment in MGA. The updated flow realigns the SPAD slices and reconstructs the representations for the next iteration. This feedback follows from the three-dimensional spatiotemporal structure of the SPAD streams. Unlike RGB inputs observed at two individual timestamps, the spatial information in the SPAD stream depends on how photon detections are aggregated over time. The SPAD representations are therefore progressively improved as the flow estimate is refined.

We also follow WAFT and supervise all $K$ predictions with the Mixture-of-Laplace negative log-likelihood. Let $\ell_k^{\mathrm{MoL}}(u)$ denote the loss of the $k$-th prediction at pixel $u$. The overall objective is
\begin{equation}
    \mathcal{L}
    =
    \sum_{k=1}^{K}\gamma^{K-k}
    \frac{1}{|\Omega|}
    \sum_{u\in\Omega}\ell_k^{\mathrm{MoL}}(u),
    \label{eq:sequence_loss}
\end{equation}
where $\Omega$ is the set of all image pixels. We set $\gamma=0.85$.

\section{Experiments}
\subsection{Experimental Setup}

\subsubsection{Dataset.}
We construct a simulated SPAD optical-flow dataset by re-rendering the assets from the 50 indoor scenes provided by VisionSIM~\cite{jungerman2025visionsim}. All sequences are rendered at 2kHz. We use 40 scenes for training and the remaining scenes for testing. Each scene contains four sequences, two under the $\Delta t=10$ setting and two under the $\Delta t=20$ setting. For every sequence, we first render the RGB frames and ground-truth optical flow frame by frame, and then sample each RGB frame into a three-channel binary photon slice. The resulting RGB, flow, and photon sequences have a spatial resolution of $512\times512$. During training, we randomly crop them to $384\times384$.

For the main benchmark, we set the illumination scale in Eq.~\eqref{eq:bernoulli} to $\alpha=0.8$. We additionally generate variants with $\alpha=0.5$ and $\alpha=0.1$ for the illumination ablation analysis. The same rendered RGB and flow sequences are used at all three illumination levels.

For real-world evaluation, we capture SPAD streams with a SPADAlpha camera~\cite{PIImagingDocs}. The raw output consists of $1024\times1024$ single-channel photon slices with BGGR Bayer pattern, recorded at 20kHz. We apply ten-times temporal downsampling to obtain a 2kHz stream and pack each raw slice into a three-channel $512\times512$ photon slice. Since the real sequences do not provide ground-truth flow, they are used for qualitative evaluation.

\subsubsection{Comparison Methods.}

The comparison methods in Table~\ref{tab:main_results} are divided into two groups. Frame-based methods first reconstruct the SPAD stream into a pair of frames and then apply the corresponding two-frame optical-flow network, whereas event/spike-based methods either convert the SPAD stream into events or directly process it as spike data. We additionally evaluate RAFT~\cite{raft2020} and WAFT~\cite{waft2026} on the ground-truth RGB frames at $t_1$ and $t_2$ as references. RAFT-r5 and WAFT-r5 integrate 11 photon slices around each timestamp to simulate short-exposure frames, while RAFT-r35 and WAFT-r35 integrate 71 slices to simulate long-exposure frames. QBP-WAFT reconstructs the images at $t_1$ and $t_2$ with Quanta Burst Photography~\cite{ma2020qbp} and then applies WAFT. For E-RAFT~\cite{gehrig2021eraft}, we convert the SPAD stream into events following SoDaCam~\cite{sundar2023sodacam}. HiST-SFlow~\cite{zhao2024hstsf}, SCFlow~\cite{hu2022scflow}, and Spike2Flow++~\cite{zhao2026scflow} directly take the SPAD stream as the binary spike stream. We report endpoint error (EPE), angular error (AE), and the fractions of pixels with EPE above 1, 2, and 3 pixels.

\subsection{Main Results}
\begin{figure*}[t]
    \centering
    \includegraphics[width=\textwidth]{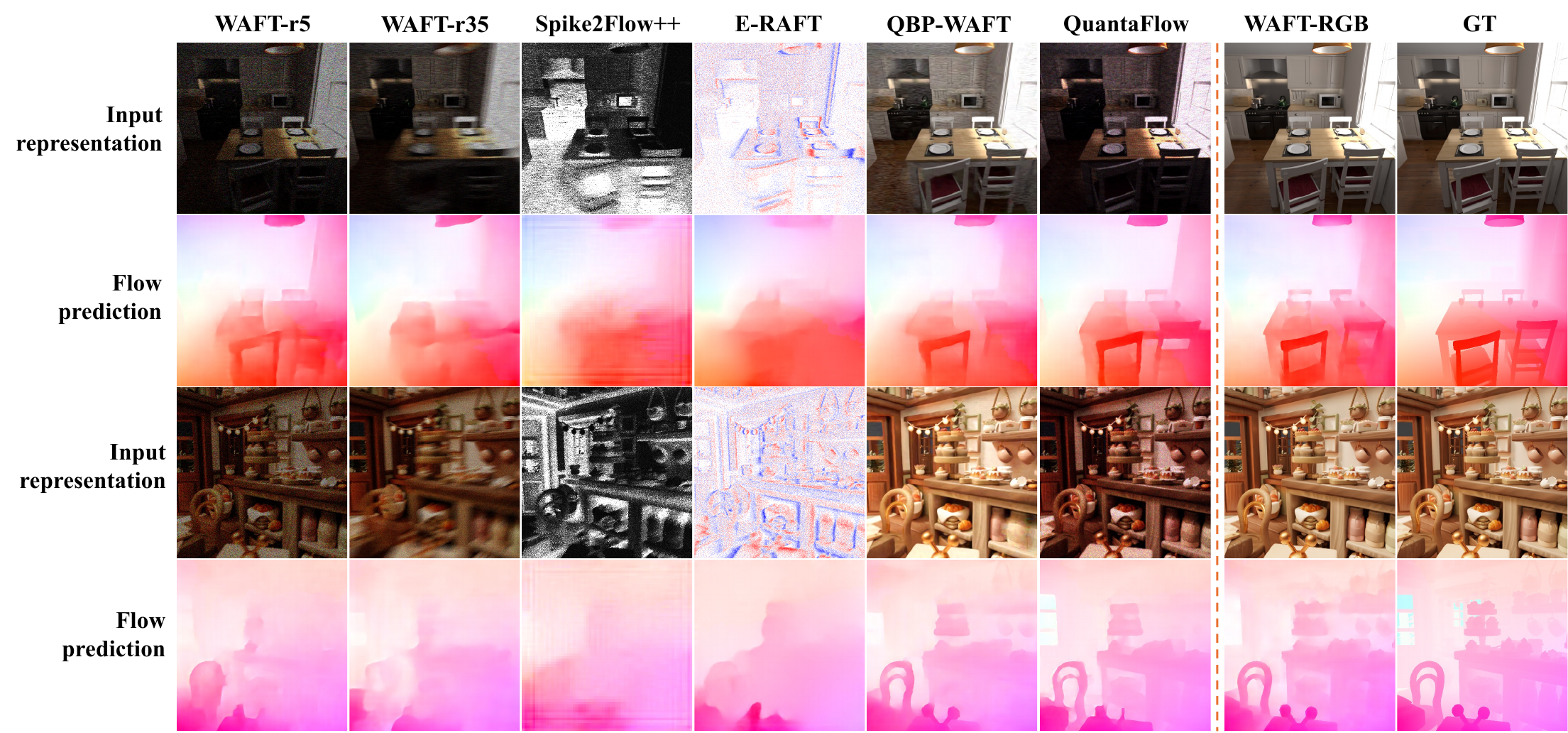}
    \caption{Comparison of input representations and optical flow on the simulated benchmark. QuantaFlow preserves spatial structures from the photon stream and produces flow boundaries that more closely match the ground truth than other methods operating on SPAD data.}
    \label{fig:qual_sim}
\end{figure*}
\begin{figure*}[t]
    \centering
    \includegraphics[width=\textwidth]{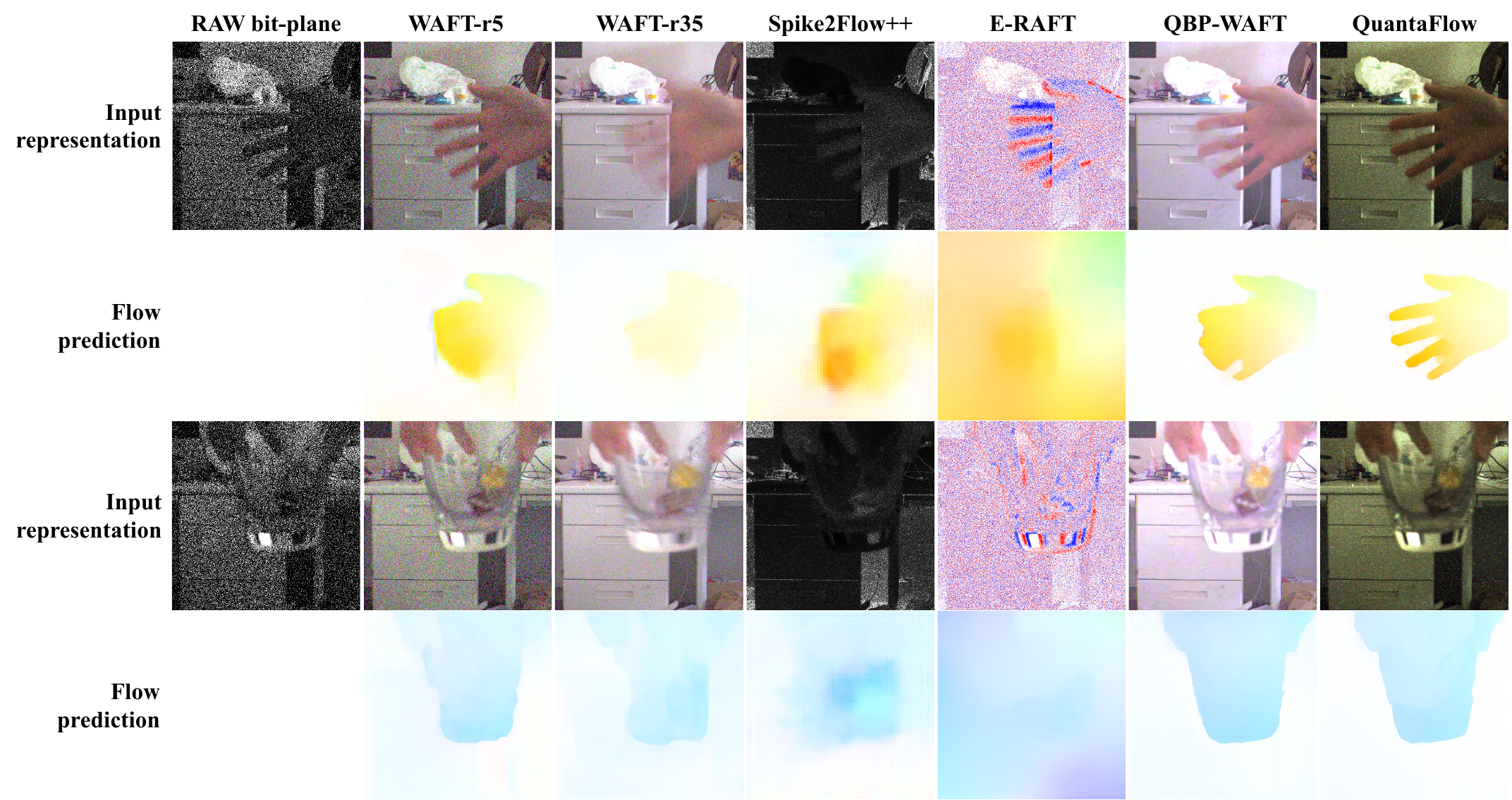}
    \caption{Qualitative comparison on real SPAD sequences. For each scene, the upper row shows the input representation and the lower row shows the predicted flow. QuantaFlow recovers coherent object motion while preserving fine motion boundaries.}
    \label{fig:qual_real}
\end{figure*}
Table~\ref{tab:main_results} reports the quantitative comparison for the $\Delta t=10$ and $\Delta t=20$ settings. QuantaFlow achieves the best result among all methods operating on SPAD data for every metric in both settings. Compared with QBP-WAFT, the strongest competing method in EPE, QuantaFlow reduces EPE by 17.3\% at $\Delta t=10$ and 2.9\% at $\Delta t=20$.

Fig.~\ref{fig:qual_sim} shows how the input representation affects the estimated flow. WAFT-r5 uses only 11 binary measurements, leaving strong photon fluctuations and fragmented structures. WAFT-r35 accumulates 71 measurements at fixed sensor coordinates, causing motion to smear object contours. Both effects lead to incomplete or poorly localized flow boundaries. Spike2Flow++ constructs a dual differential of spike firing time (DSFT) representation under the integrate-and-fire model of spike cameras. In a SPAD stream, however, the interval between detections is a stochastic photon-arrival gap rather than the time required to reach a firing threshold, making the resulting intensity cues unstable. E-RAFT re-encodes the detections as brightness changes, producing edge-dominated representations and less coherent flow fields.

QBP-WAFT removes most photon noise and produces visually clean images, but its independently optimized reconstruction smooths weak structures and fine boundaries, and remains fixed during flow estimation. QuantaFlow aligns photon detections with the current flow and rebuilds a flow-optimized representation after each update. It consequently preserves clearer object contours and produces flow closer to the ground truth.

\subsection{Ablation Studies}

\paragraph{Module ablation.}
Table~\ref{tab:ablation_components} evaluates the contribution of each module in QuantaFlow. The four variants are defined as follows:
\begin{itemize}
    \item \textbf{w/o MGA} accumulates SPAD slices at fixed sensor coordinates without motion-guided alignment.
    \item \textbf{w/o PFT} directly uses the normalized photon accumulation $p_{t_e}^{k,r_i}$ as the representation at each scale.
    \item \textbf{w/o AMF} replaces the adaptive fusion weights with a uniform average across scales.
    \item \textbf{w/o IRU} keeps the initially constructed SPAD representations fixed throughout flow refinement.
\end{itemize}
Each variant changes only the specified component and is trained with the same configuration as the full model.
\begin{table}[h]
\centering
\resizebox{\columnwidth}{!}{%
\begin{tabular}{lcccc}
\toprule
Variant & \multicolumn{2}{c}{$\Delta t=10$} & \multicolumn{2}{c}{$\Delta t=20$} \\
\cmidrule(lr){2-3}\cmidrule(lr){4-5}
 & EPE $\downarrow$ & AE $\downarrow$ & EPE $\downarrow$ & AE $\downarrow$ \\
\midrule
w/o MGA & 1.3385 & 6.0524 & 2.9645 & 5.4880 \\
w/o PFT & 1.1674 & 5.5221 & 2.6458 & 5.0207 \\
w/o AMF & 1.1683 & 5.4443 & 2.6285 & 5.0349 \\
w/o IRU & 1.3681 & 6.0396 & 2.9863 & 5.4833 \\
QuantaFlow (full) & \textbf{1.1502} & \textbf{5.3361} & \textbf{2.6096} & \textbf{4.9139} \\
\bottomrule
\end{tabular}
}
\caption{Ablation study of QuantaFlow modules.}
\label{tab:ablation_components}
\end{table}

All four modules improve performance under both $\Delta t$ settings, especially MGA and IRU. Without motion-guided intra-window alignment, detections from different times are accumulated at mismatched spatial locations, degrading the representation with motion blur. Without iterative representation update, the SPAD representation cannot be progressively refined as the flow estimate improves. These results show that intra-window alignment and iterative updating are particularly important for improving the quality of the SPAD representation.

\paragraph{Illumination ablation.}
Table~\ref{tab:photon_budget} evaluates QuantaFlow and representative frame-, spike-, and event-based methods at $\alpha=0.8$, $0.5$, and $0.1$. Each method is trained and evaluated separately at every illumination level.
\begin{table}[h]
\centering
\resizebox{\columnwidth}{!}{%
\begin{tabular}{lcccccc}
\toprule
Method & \multicolumn{2}{c}{$\alpha=0.8$} & \multicolumn{2}{c}{$\alpha=0.5$} & \multicolumn{2}{c}{$\alpha=0.1$} \\
\cmidrule(lr){2-3}\cmidrule(lr){4-5}\cmidrule(lr){6-7}
 & $\Delta t=10$ & $\Delta t=20$ & $\Delta t=10$ & $\Delta t=20$ & $\Delta t=10$ & $\Delta t=20$ \\
\midrule
QBP-WAFT & 1.3902 & 2.6882 & 1.3645 & 2.7131 & 1.9512 & 3.9423 \\
Spike2Flow++ & 2.2212 & 4.5852 & 2.4290 & 4.9765 & 2.6512 & 5.4529 \\
E-RAFT & 1.9351 & 4.2389 & 2.3701 & 4.3970 & 3.6936 & 6.3136 \\
\midrule
QuantaFlow & \textbf{1.1502} & \textbf{2.6096} & \textbf{1.2558} & \textbf{2.7012} & \textbf{1.4898} & \textbf{3.1223} \\
\bottomrule
\end{tabular}
}
\caption{Ablation study of illumination scales $\alpha$. EPE results are reported for the $\Delta t=10$ and $\Delta t=20$ settings.}
\label{tab:photon_budget}
\end{table}

QuantaFlow achieves the lowest EPE at all three illumination levels. As $\alpha$ decreases, sparse detections make the event conversion of E-RAFT and the firing-interval representation of Spike2Flow++ increasingly unreliable. QBP-WAFT remains competitive through image reconstruction, but loses weak structures when too few photons are available. At $\alpha=0.1$, QuantaFlow reduces the mean EPE across the two $\Delta t$ settings by 21.7\% relative to QBP-WAFT. Under low illumination, AMF increases the contribution of representations constructed from longer temporal windows, retaining sufficient photon evidence for reliable flow estimation. Additional results are provided in the supplementary material.

\subsection{Results on Real Scenes}
Since ground-truth optical flow is unavailable for the real-world data, we qualitatively compare all methods on two real SPAD sequences in Fig.~\ref{fig:qual_real}. WAFT-r5 retains strong photon noise, while WAFT-r35 blurs moving structures through longer fixed accumulation, leading to incomplete or oversmoothed hand motion. Spike2Flow++ produces blocky responses, and E-RAFT spreads motion into the background. QBP-WAFT reconstructs clear frames and captures the dominant motion, but merges fine structures between the fingers. QuantaFlow preserves the separation of the fingers and produces a sharper hand boundary. It also recovers a complete and continuous motion region for the moving object in the second sequence. These results show that QuantaFlow achieves better generalization to real SPAD data.

\section{Conclusion}
We presented QuantaFlow, the first method for dense optical flow estimation directly from SPAD photon streams. QuantaFlow embeds SPAD representation construction into iterative flow refinement to extract stable spatial information from sparse binary observations. At each iteration, motion-guided alignment coarsely aligns photon slices with the current flow, reducing blur before aggregation. Photon-flux transformation constructs multi-scale intensity and spatial representations, while adaptive fusion combines them according to local photon noise and motion. The fused representations update the flow through feature-level warping, and the refined flow improves slice alignment in the next iteration. We constructed a synthetic dataset, and experiments on synthetic and real-world SPAD data demonstrate the effectiveness and generalization of QuantaFlow.
\subsubsection{Limitation.}
Our method relies on the assumption that displacements vary approximately uniformly within a short temporal window, performing well on high-speed macroscopic motions such as rapid rigid translation. However, it may fail under high-speed rotation, frequent occlusions, and other complex motions. Extending the intra-window motion model to such cases is left for future work.

\bibliography{aaai2027}

% \clearpage
\clearpage
% Supplementary material (inlined)
\makeatletter
\twocolumn[
  \vbox to \titlebox{%
    \hsize\textwidth%
    \linewidth\hsize%
    \vskip 0.625in minus 0.125in%
    \centering%
    {\LARGE\bf Optical Flow from Photons \\ Supplementary Material\par}%
    \vskip 1em plus 2fil%
  }%
]
\makeatother

% Supplementary material starts here.

\setcounter{section}{0}
\setcounter{subsection}{0}
\setcounter{figure}{0}
\setcounter{table}{0}
\setcounter{secnumdepth}{2}
\renewcommand{\thesection}{\Alph{section}}
\renewcommand{\thesubsection}{\thesection.\arabic{subsection}}
\renewcommand{\thefigure}{S\arabic{figure}}
\renewcommand{\thetable}{S\arabic{table}}

\section{Analysis of QuantaFlow Modules}

\subsection{Motion-Guided Alignment}
\begin{figure}[h]
  \centering
  \includegraphics[width=\columnwidth]{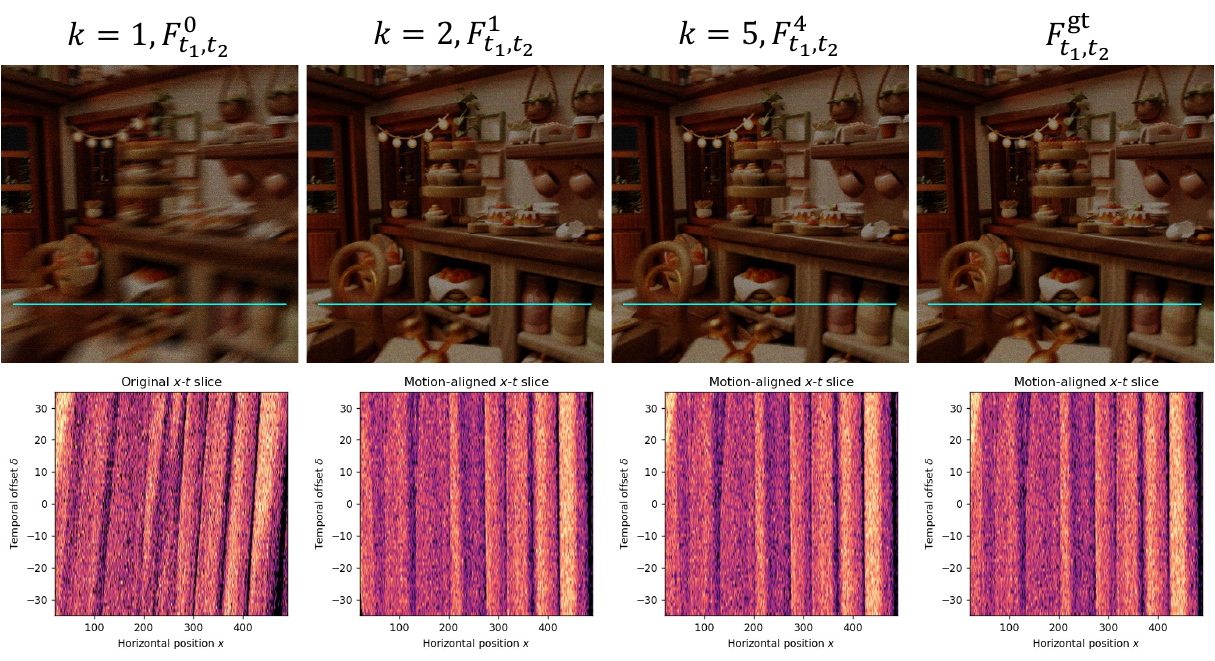}
  \caption{Effect of motion-guided alignment at the longest temporal scale $r_i=35$. The top row shows the normalized photon accumulations constructed with $F^0_{t_1,t_2}$, $F^1_{t_1,t_2}$, $F^4_{t_1,t_2}$, and the ground-truth flow. The bottom row shows the corresponding $x$--$t$ slices sampled along the cyan scan line.}
  \label{fig:motion-guided-alignment}
\end{figure}

We further visualize how MGA affects the aggregation of spatiotemporal information in the SPAD photon stream, as shown in Fig.~\ref{fig:motion-guided-alignment}. With the initial flow $F^0_{t_1,t_2}=0$, photons are accumulated at fixed sensor coordinates. Moving structures are consequently smeared in the accumulated representation, while their trajectories appear inclined and dispersed in the $x$--$t$ slice. As the flow estimate is refined, MGA increasingly follows the motion trajectories of the scene points. The accumulation constructed with $F^4_{t_1,t_2}$ exhibits sharper spatial structures, and its $x$--$t$ trajectories become more vertically concentrated, approaching the alignment obtained with the ground-truth flow. This progression shows that the refined flow improves photon aggregation in the next iteration and provides more coherent spatial information for subsequent flow estimation.
\subsection{Photon-Flux Transformation}
\begin{figure}[h]
  \centering
  \includegraphics[width=\columnwidth]{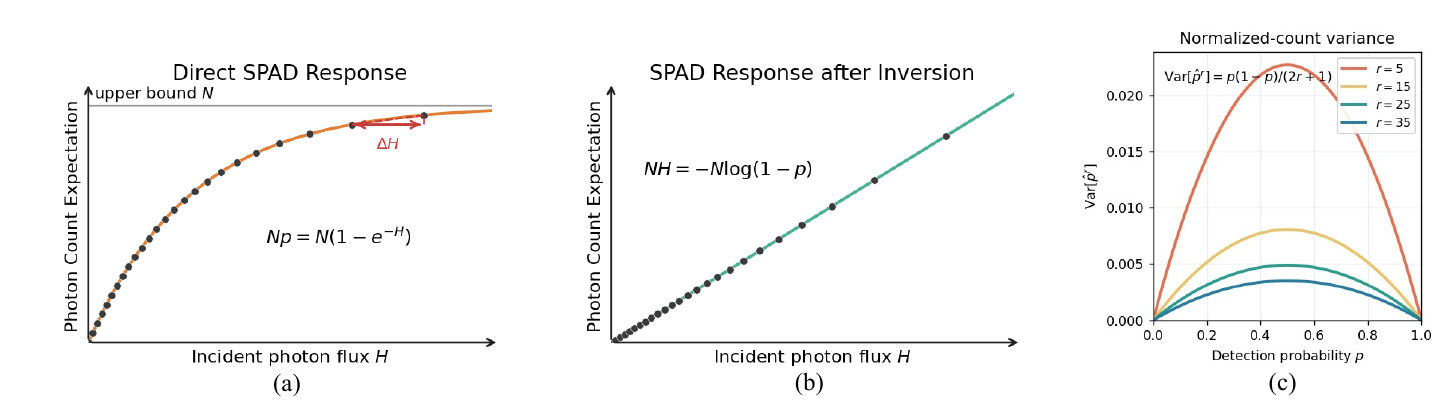}
  \caption{Statistical motivation for photon-flux transformation. (a) The expected binary photon count approaches saturation as the incident photon flux increases. (b) Applying the inverse response $H=-\log(1-p)$ recovers a linear flux response. (c) The variance of the normalized accumulation $p$ decreases as the temporal scale $r_i$ increases.}
  \label{fig:photon-flux-transformation}
\end{figure}

\begin{figure}[h]
  \centering
  \includegraphics[width=\columnwidth]{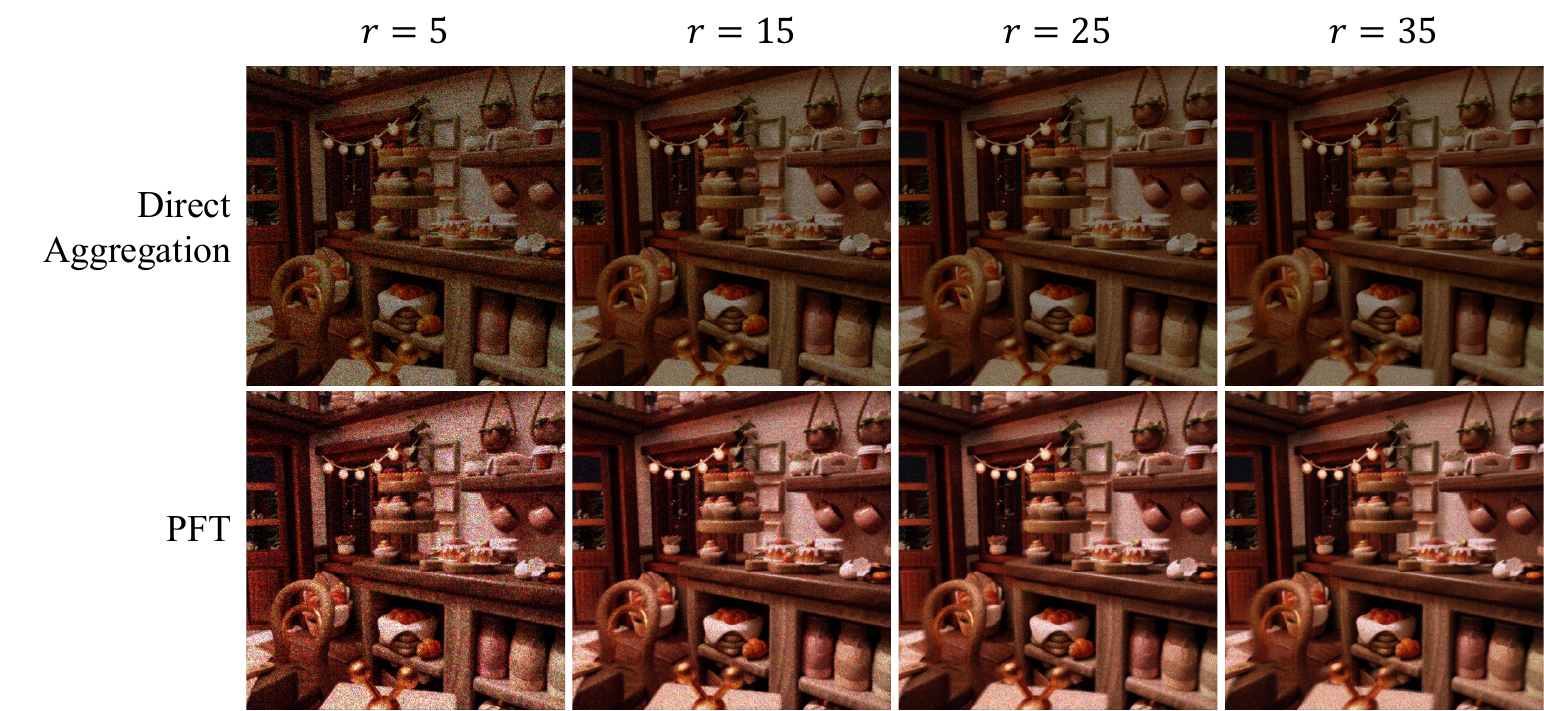}
  \caption{Effect of photon-flux transformation at different temporal scales. The top row shows the normalized photon accumulations $p^{k,r_i}_{t_1}$ for $r_i=5,15,25,35$, and the bottom row shows the corresponding scale-specific SPAD representations $f^{k,r_i}_{t_1}$ produced by PFT.}
  \label{fig:photon-flux-transformation-2}
\end{figure}

Fig.~\ref{fig:photon-flux-transformation} further illustrates the statistical motivation for PFT. For $2r_i+1$ binary observations, the expected detection count follows $(2r_i+1)p=(2r_i+1)(1-e^{-H})$ and gradually approaches its upper bound as the incident photon flux increases. Near saturation, the same change in the detection count corresponds to a larger change in the inferred flux, making direct inversion more sensitive to photon fluctuations. The transformation $H=-\log(1-p)$ linearizes the mean response, but the reliability of the estimate remains dependent on the temporal scale. As shown in Fig.~\ref{fig:photon-flux-transformation}(c), the variance of the normalized accumulation decreases with $r_i$, so short-scale estimates exhibit substantially stronger statistical fluctuations. The detection probability $p$, recovered flux $H$, and temporal scale $r_i$ therefore provide complementary information for evaluating the reliability of the SPAD observation.
Fig.~\ref{fig:photon-flux-transformation-2} visualizes the resulting representations at the four temporal scales. The shortest scale retains sharp spatial changes but contains pronounced photon fluctuations, whereas longer scales provide increasingly stable intensity and structural cues. PFT jointly interprets $p$, $H$, and $r_i$ to produce scale-specific SPAD representations with clearer structures and more consistent responses across scales. These representations preserve the distinct information provided by each temporal scale and form the candidates subsequently combined by AMF.

\subsection{Adaptive Multi-Scale Fusion}
\begin{figure}[h]
  \centering
  \includegraphics[width=\columnwidth]{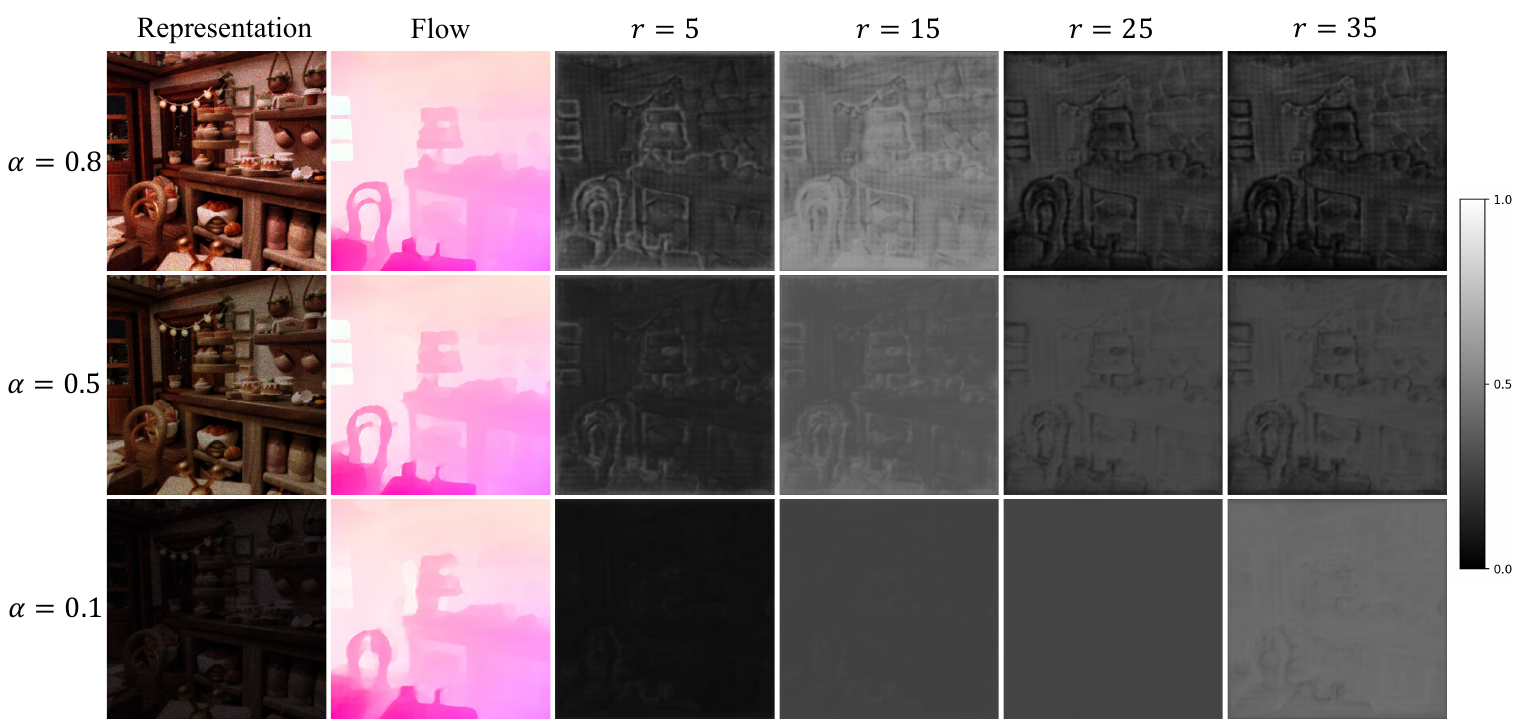}
  \caption{Adaptive multi-scale fusion under different illumination scales. Each row shows the fused SPAD representation, flow estimate, and pixel-wise weights for $r_i=5,15,25,35$ at the final iteration. Brighter values indicate larger weights. As $\alpha$ decreases, AMF progressively shifts its weights toward longer temporal scales.}
  \label{fig:supp_AMF}
\end{figure}

Fig.~\ref{fig:supp_AMF} visualizes how AMF adapts its scale weights to the illumination level. At $\alpha=0.8$, the intermediate scale $r_i=15$ receives the largest contribution, while the shorter-scale weights retain clear spatial variations around scene structures. As $\alpha$ decreases, sparse photon detections reduce the reliability of these short-scale representations, and AMF progressively increases the contribution of $r_i=25$ and $r_i=35$. At $\alpha=0.1$, the longest scale becomes dominant and supplies sufficient photon information to maintain a coherent fused representation and flow estimate. Over the complete evaluation set, the mean expected temporal scale increases from $16.00$ at $\alpha=0.8$ to $21.16$ at $\alpha=0.5$ and $25.51$ at $\alpha=0.1$. These results show that AMF adjusts the amount of temporal information according to the available photon observations.

\subsection{Iterative Representation Update}
\begin{figure}[h]
  \centering
  \includegraphics[width=\columnwidth]{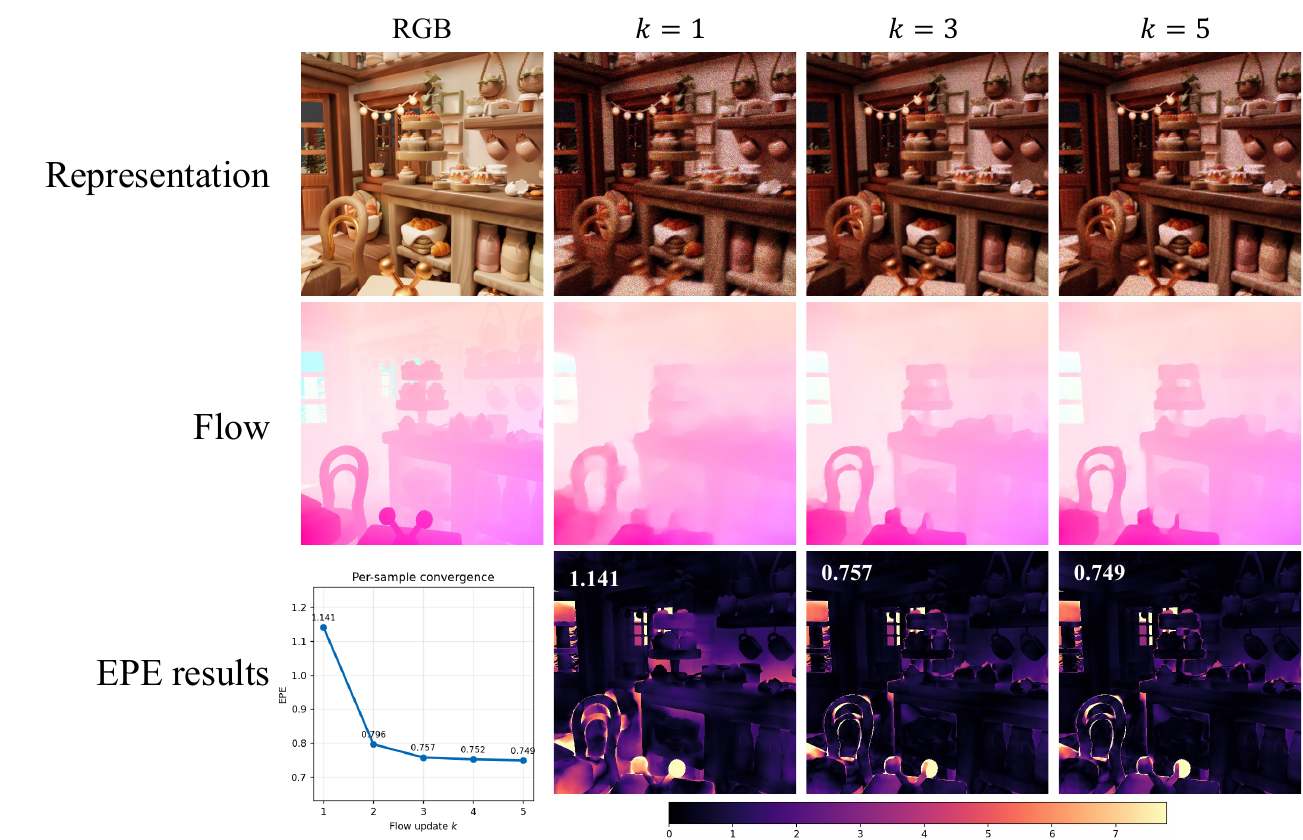}
  \caption{Improvement of the SPAD representation and flow estimate through iterative representation update. The three columns show the representations, flow estimates, and endpoint-error maps after the first, third, and fifth updates. The RGB reference, ground-truth flow, and per-sample EPE convergence curve are shown in the first column.}
  \label{fig:iterative-representation-update}
\end{figure}

Fig.~\ref{fig:iterative-representation-update} shows how the SPAD representation and flow estimate evolve through IRU. The first representation is constructed with the zero initialization $F^0_{t_1,t_2}$ and produces a coarse flow estimate with an EPE of $1.141$. This flow guides MGA in the next iteration, allowing the photon detections to be aggregated along more accurate motion trajectories. The newly constructed representation consequently contains sharper spatial structures and supports a more accurate flow update. The EPE decreases to $0.757$ after the third update and further converges to $0.749$ after the fifth update. The simultaneous improvement of the representation and flow illustrates the feedback between motion-guided SPAD representation construction and iterative flow refinement.

\section{Dataset Construction}

\subsection{Synthetic Dataset Generation}
Table~\ref{tab:supp_dataset_config} summarizes the dataset split and sample counts. Each scene contains two sequences for $\Delta t=10$ and two sequences for $\Delta t=20$. The two settings contain 285 and 315 rendered frames per sequence, respectively, and each sequence provides three flow pairs. The 50 scenes are divided into 40 training scenes and 10 test scenes without scene overlap.

Indexing the slices in each sequence from zero, the central frames for $\Delta t=10$ are $127$, $137$, $147$, and $157$. The three ground-truth forward flows correspond to the frame pairs $(127,137)$, $(137,147)$, and $(147,157)$. For $\Delta t=20$, the central frames are $127$, $147$, $167$, and $187$, with ground-truth flows for $(127,147)$, $(147,167)$, and $(167,187)$. This sequence layout supports temporal aggregation with a maximum scale of $r=127$ around both frames of every pair. In our experiments, the additional photon detections collected by longer windows were outweighed by the resulting motion blur, which reduced flow accuracy. We therefore set the largest temporal scale to $r=35$ for both training and evaluation.

\begin{table}[h]
  \centering
  \small
  \begin{tabular}{lcc}
    \toprule
    Configuration & Training & Test \\
    \midrule
    Scenes & 40 & 10 \\
    Sequences & 160 & 38 \\
    Frame rate & 2 kHz & 2 kHz \\
    Spatial resolution & $512\times512$ & $512\times512$ \\
    Channels & 3 & 3 \\
    Frames/sequence ($\Delta t=10$) & 285 & 285 \\
    Frames/sequence ($\Delta t=20$) & 315 & 315 \\
    Total frames & 48,000 & 11400 \\
    Flow pairs ($\Delta t=10$) & 240 & 57 \\
    Flow pairs ($\Delta t=20$) & 240 & 57 \\
    Total flow pairs & 480 & 114 \\
    \bottomrule
  \end{tabular}
  \caption{Configuration of the synthetic SPAD optical-flow dataset.}
  \label{tab:supp_dataset_config}
\end{table}

\subsection{Real-World SPAD Data Acquisition}
We record several real-world scenes using a SPADAlpha camera. Each acquisition contains 1,000 raw $1024\times1024$ binary slices captured at 20 kHz with a BGGR Bayer pattern. To match the 2 kHz frame rate of the synthetic data, we retain one slice from every ten consecutive slices, corresponding to raw indices $0,10,20,\ldots,990$. This operation yields a sequence of 100 slices and is equivalent to reducing the effective camera readout rate from 20 kHz to 2 kHz. Each retained slice remains an individual exposure from the original stream, so its exposure time and per-slice photon-detection probability are unchanged. We then pack each raw Bayer slice into a three-channel $512\times512$ photon slice. For qualitative evaluation, the frame pairs are $(45,55)$ for $\Delta t=10$ and $(40,60)$ for $\Delta t=20$, both of which support the temporal scales used by QuantaFlow.

\section{Implementation Details}

\subsection{Network and Training Configuration}
The PFT estimator is shared across the four temporal scales and between the source and target sub-streams. Its input contains the three-channel detection probability $p^{k,r_i}_{t_e}$, the three-channel recovered flux $H^{k,r_i}_{t_e}$, and a two-channel encoding of the temporal scale. As detailed in Table~\ref{tab:supp_pft_architecture}, two convolutional layers transform these statistics before a residual head predicts a three-channel correction to the recovered flux. The residual scale is set to $0.25$.

\begin{table}[t]
  \centering
  \small
  \setlength{\tabcolsep}{4pt}
  \begin{tabular}{lccc}
    \toprule
    Layer & Kernel & In ch. & Out ch. \\
    \midrule
    Conv + GN(8) + GELU & $3\times3$ & 8 & 32 \\
    Conv + GN(8) + GELU & $3\times3$ & 32 & 32 \\
    Conv + Tanh & $3\times3$ & 32 & 3 \\
    \bottomrule
  \end{tabular}
  \caption{Network structure of the photon-flux transformation estimator. GN(8) denotes group normalization with eight groups.}
  \label{tab:supp_pft_architecture}
\end{table}

AMF predicts four pixel-wise scale weights using the motion magnitude associated with each scale, the flux and scale statistics produced during PFT, and the recurrent context from the preceding update. The 64-channel recurrent feature is first compressed to eight channels. Concatenating it with the four-channel motion descriptor, four-channel flux descriptor, and four-channel scale descriptor gives the 20-channel input listed in Table~\ref{tab:supp_amf_architecture}. A softmax with temperature $2.0$ converts the predicted logits into normalized scale weights.

\begin{table}[t]
  \centering
  \small
  \setlength{\tabcolsep}{4pt}
  \begin{tabular}{lccc}
    \toprule
    Layer & Kernel & In ch. & Out ch. \\
    \midrule
    Context Conv & $1\times1$ & 64 & 8 \\
    Conv + GELU & $3\times3$ & 20 & 64 \\
    Conv + GELU & $3\times3$ & 64 & 64 \\
    Weight Conv & $1\times1$ & 64 & 4 \\
    Softmax & -- & 4 & 4 \\
    \bottomrule
  \end{tabular}
  \caption{Network structure of the adaptive multi-scale fusion module.}
  \label{tab:supp_amf_architecture}
\end{table}

For feature extraction, we adopt the encoder used by WAFT. It combines a frozen ImageNet-pretrained Twins-SVT-Large backbone with trainable multi-scale feature-fusion layers and a parallel convolutional branch. Their outputs are projected to a 64-channel feature map at half resolution. The recurrent flow-update unit uses the ViT-S configuration with a DPT decoder.

The WAFT encoder and update unit are initialized from the official Twins zero-shot checkpoint, while PFT and AMF are initialized from scratch. The final convolution of PFT is initialized to zero, so the initial residual vanishes and the module starts from the recovered flux $H$. The AMF weight head is also initialized with zero weights and biases, giving an initial weight of $1/4$ to each temporal scale. All trainable parameters are then optimized jointly using the settings in Table~\ref{tab:supp_training_config}.

\begin{table}[h]
  \centering
  \small
  \begin{tabular}{lc}
    \toprule
    Configuration & Setting \\
    \midrule
    Training steps & 100,000 \\
    Batch size & 4 \\
    Optimizer & AdamW \\
    Peak learning rate & $2\times10^{-5}$ \\
    Learning-rate schedule & One-cycle, linear \\
    Warm-up & First 5\% of training \\
    Weight decay & $5\times10^{-5}$ \\
    Optimizer $\epsilon$ & $10^{-8}$ \\
    Gradient clipping & 1.0 \\
    Flow-update iterations & 5 \\
    Sequence-loss decay $\gamma$ & 0.85 \\
    Training crop & $384\times384$ \\
    Rotation augmentation & $0^\circ,90^\circ,180^\circ,270^\circ$ \\
    Training intervals & Alternating $\Delta t=10/20$ \\
    \bottomrule
  \end{tabular}
  \caption{Training configuration of QuantaFlow.}
  \label{tab:supp_training_config}
\end{table}

\begin{table*}[t]
  \centering
  \resizebox{\textwidth}{!}{%
  \begin{tabular}{lcccccccccc}
    \toprule
    Variant & \multicolumn{5}{c}{$\Delta t=10$} & \multicolumn{5}{c}{$\Delta t=20$} \\
    \cmidrule(lr){2-6}\cmidrule(lr){7-11}
     & EPE $\downarrow$ & AE $\downarrow$ & 1PE $\downarrow$ & 2PE $\downarrow$ & 3PE $\downarrow$
     & EPE $\downarrow$ & AE $\downarrow$ & 1PE $\downarrow$ & 2PE $\downarrow$ & 3PE $\downarrow$ \\
    \midrule
    w/o MGA & 1.3385 & 6.0524 & 0.3339 & 0.1705 & 0.1128 & 2.9645 & 5.4880 & 0.5365 & 0.3325 & 0.2314 \\
    w/o PFT & 1.1674 & 5.5221 & 0.2831 & 0.1444 & 0.0946 & 2.6458 & 5.0207 & 0.4279 & 0.2731 & 0.2015 \\
    w/o AMF & 1.1683 & 5.4443 & \textbf{0.2708} & \textbf{0.1438} & 0.0954 & 2.6285 & 5.0349 & \textbf{0.4270} & \textbf{0.2700} & \textbf{0.1996} \\
    w/o IRU & 1.3681 & 6.0396 & 0.3530 & 0.1795 & 0.1157 & 2.9863 & 5.4833 & 0.5495 & 0.3483 & 0.2460 \\
    QuantaFlow (full) & \textbf{1.1502} & \textbf{5.3361} & 0.2867 & 0.1515 & \textbf{0.0940} & \textbf{2.6096} & \textbf{4.9139} & 0.4344 & 0.2830 & 0.2111 \\
    \bottomrule
  \end{tabular}%
  }
  \caption{Complete module-ablation results under the $\Delta t=10$ and $\Delta t=20$ settings.}
  \label{tab:supp_component_ablation}

  \vspace{1ex}
  \centering
  \resizebox{\textwidth}{!}{%
  \begin{tabular}{lcccccccccc}
    \toprule
    Method & \multicolumn{5}{c}{$\Delta t=10$} & \multicolumn{5}{c}{$\Delta t=20$} \\
    \cmidrule(lr){2-6}\cmidrule(lr){7-11}
     & EPE $\downarrow$ & AE $\downarrow$ & 1PE $\downarrow$ & 2PE $\downarrow$ & 3PE $\downarrow$
     & EPE $\downarrow$ & AE $\downarrow$ & 1PE $\downarrow$ & 2PE $\downarrow$ & 3PE $\downarrow$ \\
    \midrule
    \multicolumn{11}{l}{$\alpha=0.8$} \\
    QBP-WAFT & 1.3902 & 7.1670 & 0.4185 & 0.1884 & 0.0973 & 2.6882 & 5.9406 & 0.5716 & 0.3443 & 0.2345 \\
    Spike2Flow++ & 2.2212 & 10.7403 & 0.5794 & 0.3290 & 0.2057 & 4.5852 & 10.5538 & 0.7892 & 0.5858 & 0.4432 \\
    E-RAFT & 1.9351 & 10.5408 & 0.5605 & 0.2857 & 0.1727 & 4.2389 & 10.7590 & 0.7716 & 0.5634 & 0.4155 \\
    QuantaFlow & \textbf{1.1502} & \textbf{5.3361} & \textbf{0.2867} & \textbf{0.1515} & \textbf{0.0940} & \textbf{2.6096} & \textbf{4.9139} & \textbf{0.4344} & \textbf{0.2830} & \textbf{0.2111} \\
    \midrule
    \multicolumn{11}{l}{$\alpha=0.5$} \\
    QBP-WAFT & 1.3645 & 7.2378 & 0.4039 & 0.1738 & 0.0930 & 2.7131 & 5.9967 & 0.5733 & 0.3567 & 0.2443 \\
    Spike2Flow++ & 2.4290 & 11.2526 & 0.6172 & 0.3750 & 0.2388 & 4.9765 & 10.8848 & 0.8196 & 0.6166 & 0.4771 \\
    E-RAFT & 2.3701 & 14.1430 & 0.6277 & 0.3556 & 0.2189 & 4.3970 & 11.1044 & 0.8018 & 0.5640 & 0.4127 \\
    QuantaFlow & \textbf{1.2558} & \textbf{5.7109} & \textbf{0.2882} & \textbf{0.1599} & \textbf{0.1072} & \textbf{2.7012} & \textbf{5.0401} & \textbf{0.4362} & \textbf{0.2781} & \textbf{0.2107} \\
    \midrule
    \multicolumn{11}{l}{$\alpha=0.1$} \\
    QBP-WAFT & 1.9512 & 9.4378 & 0.5677 & 0.2959 & 0.1704 & 3.9423 & 8.8915 & 0.7487 & 0.5295 & 0.3831 \\
    Spike2Flow++ & 2.6512 & 13.1754 & 0.6728 & 0.4084 & 0.2548 & 5.4529 & 13.2525 & 0.8691 & 0.6759 & 0.5263 \\
    E-RAFT & 3.6936 & 22.0783 & 0.7476 & 0.4870 & 0.3432 & 6.3136 & 16.3269 & 0.8717 & 0.6895 & 0.5472 \\
    QuantaFlow & \textbf{1.4898} & \textbf{7.0198} & \textbf{0.3664} & \textbf{0.1976} & \textbf{0.1270} & \textbf{3.1223} & \textbf{6.8325} & \textbf{0.5798} & \textbf{0.3729} & \textbf{0.2698} \\
    \bottomrule
  \end{tabular}%
  }
  \caption{Complete illumination-ablation results at $\alpha=0.8$, $0.5$, and $0.1$ under the $\Delta t=10$ and $\Delta t=20$ settings.}
  \label{tab:supp_illumination_ablation}

  \vspace{1ex}
  \includegraphics[width=0.9\textwidth]{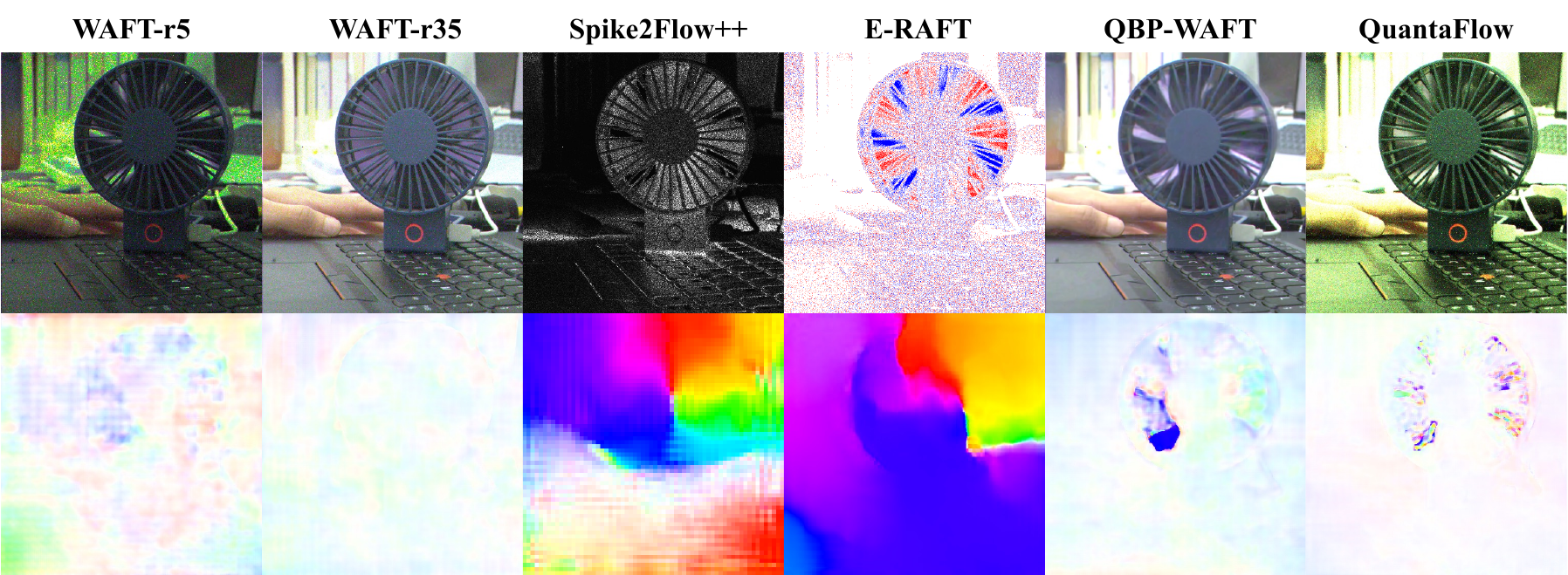}
  {\captionof{figure}{A failure case of QuantaFlow on the real-world SPAD data. Optical flow is estimated from the SPAD stream at its native frame rate of 20~kHz.}
  \label{fig:supp_fail}}
\end{table*}

\subsection{Comparison Method Configuration}
We use the released implementations of all comparison methods and retain their original network architectures. The models are trained or fine-tuned on the same synthetic training scenes and evaluated at $512\times512$ resolution. RAFT and WAFT are initialized from their published checkpoints. Their r5 and r35 variants form each input frame by accumulating 11 and 71 SPAD slices, respectively. E-RAFT uses the SPAD-to-event conversion described in the main paper, while HiST-SFlow, SCFlow, and Spike2Flow++ follow their original spike representations and process each color channel in the same manner.

For QBP-WAFT, we follow the QBP pipeline with coarse-to-fine patch alignment, Wiener merging, and photon-response inversion. Each reconstruction uses 71 slices centered at $t_e$, corresponding to $r=35$. Its maximum visible temporal window is therefore identical to that of QuantaFlow. The 71 slices are divided into seven symmetric blocks of lengths $[10,10,10,11,10,10,10]$ for alignment and merging. The reconstructed frame pair is then used to fine-tune WAFT from its zero-shot checkpoint, while the remaining QBP and WAFT settings follow their default configurations.

\section{Additional Experimental Results}

\subsection{Complete Ablation Results}
Tables~\ref{tab:supp_component_ablation} and~\ref{tab:supp_illumination_ablation} provide the complete results for the module-ablation and illumination-ablation experiments, respectively.
\subsection{Additional Qualitative Results and Limitations}
\begin{figure*}[h]
  \centering
  \includegraphics[width=\textwidth]{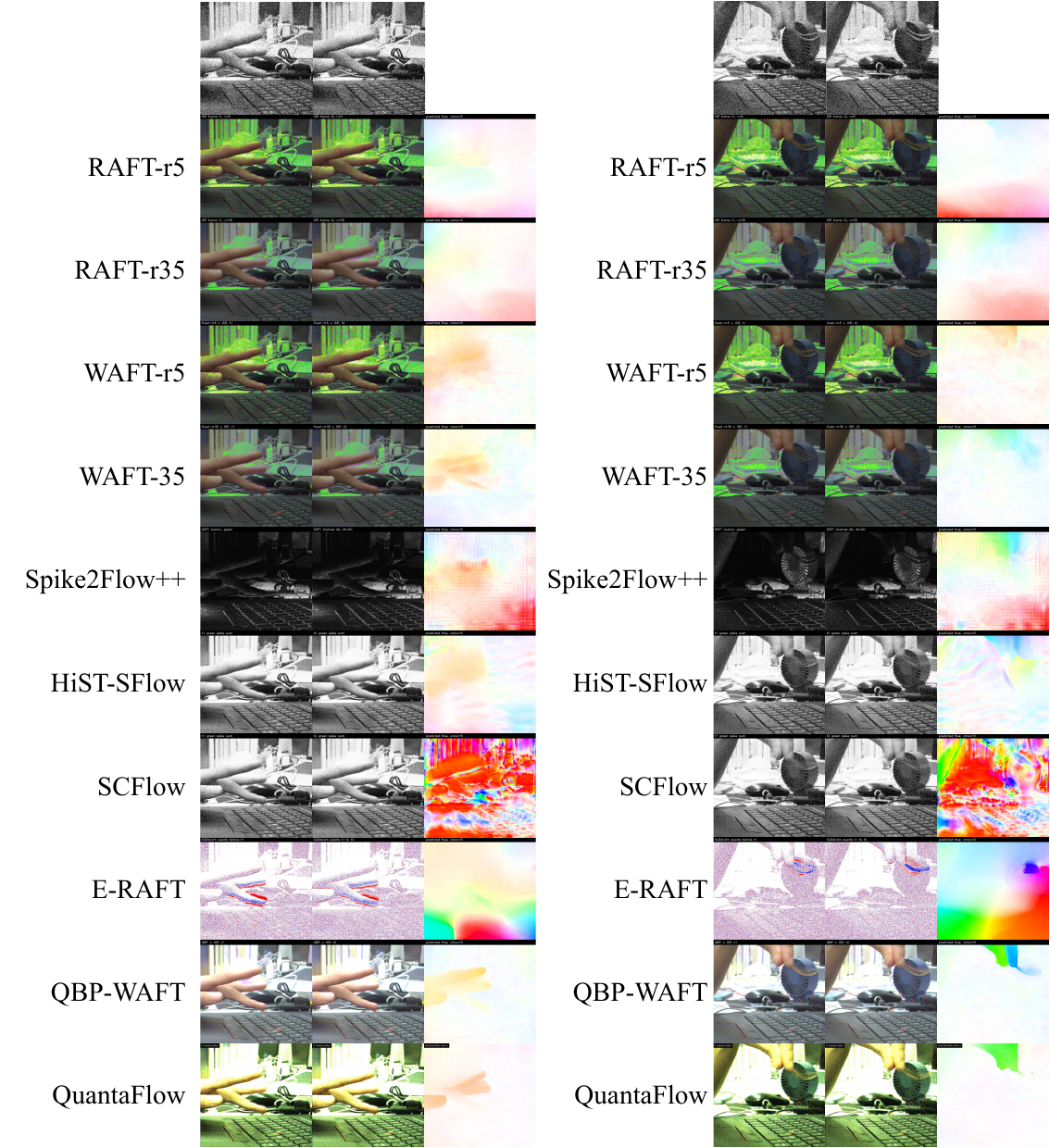}
  \caption{Additional qualitative results on the real-world SPAD data.}
  \label{fig:supp_real}
\end{figure*}

Figs.~\ref{fig:supp_fail} and~\ref{fig:supp_real} provide additional qualitative results on real-world scenes. For the rotating fan in Fig.~\ref{fig:supp_fail}, sampling the stream at 2~kHz does not preserve sufficient temporal information to resolve the trajectories of the rapidly moving blades, causing all evaluated methods to fail. We therefore perform optical flow estimation on this scene using the original 20~kHz SPAD stream.

QuantaFlow assumes that the variations of scene-point displacement between adjacent slices are stable and uniform within a short temporal window. This model is well suited to macroscopic motions with locally coherent trajectories, such as rapid rigid translation. More complex motions violate this assumption. The high-speed rotation and frequent occlusions in Fig.~\ref{fig:supp_fail}, as well as the chaotic elastic deformation of the rubber band on the right of Fig.~\ref{fig:supp_real}, cannot be described reliably by temporally scaling a single endpoint flow. These cases consequently remain difficult for QuantaFlow. Future work will extend the intra-window motion model to complex nonuniform and nonrigid motion with occlusions.

\end{document}